\documentclass{article}
\usepackage[preprint]{colm2026_conference}

\usepackage{microtype}
\usepackage{hyperref}
\usepackage{url}
\usepackage{booktabs}
\usepackage{graphicx}
\usepackage{amsmath}
\usepackage{amssymb}
\usepackage{multirow}
\usepackage{xcolor}
\usepackage{float}
\usepackage{lineno}

\definecolor{darkblue}{rgb}{0, 0, 0.5}
\hypersetup{colorlinks=true, citecolor=darkblue, linkcolor=darkblue, urlcolor=darkblue}

\title{The Randomness Floor: Measuring Intrinsic Non-Randomness in Language Model Token Distributions}

\author{Jaros{\l}aw Hryszko\thanks{ORCID: 0000-0002-4207-1080} \\
Institute of Computer Science\\
Faculty of Mathematics and Computer Science\\
Jagiellonian University, Krak{\'o}w, Poland\\
\texttt{jaroslaw.hryszko@uj.edu.pl}
}

\newcommand{\ED}{\mathrm{ED}}
\newcommand{\DKL}{D_{\mathrm{KL}}}

\begin{document}

\maketitle

\begin{abstract}
Language models cannot be random. This paper introduces Entropic Deviation (ED), the normalized KL divergence between a model's token distribution and the uniform distribution, and measures it systematically across 31,200 generations spanning seven models, two architectures (transformer and state space), nine prompt categories, three temperatures, and five languages. Under semantically neutral prompts (empty strings, random characters, nonsense syllables) transformers still exhibit ED of approximately 0.30, meaning that 88--93\% of the non-randomness observed under semantic prompts is intrinsic to the learned weights rather than induced by context. Three transformer families (Gemma, Llama, Qwen) converge on nearly identical ED values despite different training data and vocabularies. A state space model (Mamba2) reveals a qualitatively different regime: twice the ED, three times lower within-sequence variance, and massive sensitivity to temperature ($r = -0.78$) where transformers are nearly immune ($r < 0.05$). Cross-lingual experiments with Qwen-32B show a stable gradient across five languages (English, Japanese, Chinese, Polish, Arabic) that does not correlate with token fertility and persists when two languages sharing an identical tokenizer subset are compared. These findings establish a structural lower bound on randomness in pretrained language models, characterize how this bound differs across architectures, and demonstrate that language itself modulates the bound independently of tokenization.
\end{abstract}

\section{Introduction}
\label{sec:intro}

Every time a language model produces a token, it first generates a probability distribution over its entire vocabulary. This distribution is the fundamental output of the model: the token that appears in the generated text is merely a sample from it. The shape of this distribution encodes everything the model has learned about language, from syntactic constraints to world knowledge.

A natural question arises: what happens to this distribution when the model has nothing meaningful to predict? If given an empty string, a sequence of random ASCII characters, or a prompt consisting of nonsense syllables, does the model produce something close to a uniform distribution over its vocabulary, or does it remain concentrated on a subset of tokens?

The answer has implications for how we understand what language models learn. A model that can approximate uniformity under meaningless input would be one whose distributional structure is primarily driven by context. A model that remains far from uniform even without context would be one whose learned weights impose a persistent structural bias on its outputs, independent of what it is asked to generate.

This paper introduces Entropic Deviation (ED), a simple metric based on the normalized KL divergence between a model's next-token distribution and the uniform distribution over its vocabulary. ED equals zero for a perfectly uniform distribution and one for a distribution concentrated on a single token. Measured across 31,200 generations spanning seven models, two architectures, nine prompt categories, three temperatures, and five languages, ED reveals a consistent pattern: language models are structurally incapable of producing uniform token distributions, even when given no meaningful input.

The central findings organize into a sequence where each result eliminates the simplest explanation of the previous one:

\begin{enumerate}
\item \textbf{ED is nonzero everywhere.} All tested models, under all conditions, produce distributions significantly different from uniform ($p < 10^{-6}$ in all cases).

\item \textbf{ED is predominantly intrinsic.} Under semantically neutral prompts, transformers retain 88--93\% of the ED observed under meaningful prompts. The semantic content of the input adds only a small increment to a large baseline.

\item \textbf{ED is architecturally stable across transformers.} Three transformer families (Gemma-2 27B, Qwen-2.5 32B, Llama-3.3 70B), trained on different data with different tokenizers, converge on ED $\approx 0.31$ under semantic prompts and $\approx 0.30$ under neutral prompts, with inter-model domain correlations of $\rho > 0.93$.

\item \textbf{ED differs qualitatively across architectures.} A state space model (Mamba2-2.7B) exhibits ED approximately twice that of transformers, with three times lower within-sequence variance and a massive negative correlation with temperature ($r = -0.78$), while transformer ED is nearly temperature-invariant ($r < 0.16$).

\item \textbf{ED is modulated by language independently of tokenization.} Across five languages, ED follows a stable gradient (EN $<$ JA $<$ ZH $<$ PL $<$ AR) that does not correlate with token fertility ($\rho = 0.10$, $p = 0.87$) and persists when comparing English and Polish, which share an identical tokenizer allocation of 104,969 tokens.
\end{enumerate}

This work differs from the existing literature on randomness in language models in a fundamental way. Prior studies ask whether language models can \emph{generate} random outputs: random numbers~\citep{hopkins2023llm_sampling, coronado2025deterministic}, samples from statistical distributions~\citep{zhao2026bad_dice}, or sequences passing randomness tests. These studies operate at the level of what the model \emph{says}. The present work operates at the level of what the model \emph{computes}: the probability distribution over the full vocabulary at each token position, before any sampling or decoding takes place. ED measures not whether a model can be instructed to act randomly, but whether the distributions it produces can be random at all.

\section{Related work}
\label{sec:related}

\paragraph{LLM randomness at the output level.}
\citet{hopkins2023llm_sampling} evaluate language models as samplers of probability distributions, finding systematic deviations from target distributions even for simple uniform sampling tasks. \citet{zhao2026bad_dice} extend this to 11 models across 15 statistical distributions, documenting consistent failures to approximate even basic distributions when asked to generate samples. \citet{coronado2025deterministic} approach the question from a psychological perspective, analyzing the ``random number generation'' behavior of LLMs and finding patterns reminiscent of human cognitive biases. All three lines of work share a common methodology: prompting models to produce random-seeming outputs and measuring the deviation. The present work inverts this approach by examining the raw logit distributions rather than sampled outputs, and by deliberately removing semantic context to isolate the model's intrinsic distributional structure.

\paragraph{Token distributions and language structure.}
The observation that natural language token frequencies follow heavy-tailed distributions is well established~\citep{zipf1949human, piantadosi2014zipf}. Language models trained on such data are expected to reflect these distributional properties in their outputs. However, the degree to which this structure persists under minimal or absent context, and whether it differs across architectures, has not been systematically characterized.

\paragraph{Architectures.}
The transformer architecture~\citep{vaswani2017attention} dominates current language modeling. State space models (SSMs) represent an alternative paradigm, with Mamba~\citep{gu2023mamba} and its successor Mamba2~\citep{dao2024mamba2} offering linear-time sequence processing through selective state space mechanisms. The architectural differences between attention-based and recurrence-based models may produce different distributional properties, but direct comparisons at the logit level have not been reported.

\paragraph{Temperature and decoding.}
Temperature scaling divides logits by a scalar before the softmax operation, controlling the entropy of the output distribution~\citep{holtzman2020curious}. Higher temperatures produce flatter distributions; lower temperatures sharpen them. The effect is well studied for text quality~\citep{guo2017calibration} but not for the baseline distributional structure under neutral inputs.

\paragraph{Multilingual models and tokenization.}
Tokenizer design affects how models process different languages, with fertility (tokens per character) varying substantially across scripts~\citep{rust2021good, sennrich2016neural}. Whether these differences in tokenization drive differences in distributional concentration, or whether language-level properties independently modulate model behavior, has not been tested with controlled experiments.

\section{Method}
\label{sec:method}

\subsection{Entropic Deviation}

For a language model with vocabulary size $V$, let $p = (p_1, \ldots, p_V)$ denote the softmax output distribution at a given token position, and let $u = (1/V, \ldots, 1/V)$ denote the uniform distribution. The Entropic Deviation is defined as:

\begin{equation}
\ED(p) = \frac{\DKL(p \| u)}{\log V} = \frac{\sum_{i=1}^{V} p_i \log \frac{p_i}{1/V}}{\log V} = 1 - \frac{H(p)}{\log V}
\label{eq:ed}
\end{equation}

\noindent where $H(p) = -\sum_i p_i \log p_i$ is the Shannon entropy and all logarithms are natural. ED equals zero when $p$ is uniform and one when $p$ places all mass on a single token. The normalization by $\log V$ makes ED comparable across models with different vocabulary sizes.

For a generated sequence of length $L$, the ED of the full generation is computed as the mean ED across all token positions:

\begin{equation}
\ED_{\text{seq}} = \frac{1}{L} \sum_{t=1}^{L} \ED(p_t)
\end{equation}

\noindent The within-sequence standard deviation $\ED_{\text{std}}$ is also recorded and provides a measure of how uniformly concentrated the model's distributions are across positions.

\subsection{Experimental design}

The experiment spans three phases: a pilot study on smaller models, a main study comparing transformers and an SSM, and a multilingual extension. Table~\ref{tab:models} summarizes the models.

\begin{table}[t]
\centering
\small
\begin{tabular}{llrrl}
\toprule
\textbf{Model} & \textbf{Arch.} & \textbf{Params} & \textbf{Vocab} & \textbf{Phase} \\
\midrule
Phi-3-mini-4K & TF & 3.8B & 32K & Pilot \\
Mistral-7B & TF & 7B & 32K & Pilot \\
Llama-3-8B & TF & 8B & 128K & Pilot \\
\midrule
Gemma-2-27B-IT & TF & 27B & 256K & Main \\
Qwen-2.5-32B-Instruct & TF & 32B & 152K & Main \\
Llama-3.3-70B-Instruct & TF & 70B & 128K & Main \\
\midrule
Mamba2-2.7B & SSM & 2.7B & 50K & Main \\
\bottomrule
\end{tabular}
\caption{Models used in the study. TF = transformer, SSM = state space model. All models were run locally using llama.cpp~\citep{llama_cpp} in GGUF format with Q4\_K\_M quantization on dual NVIDIA Tesla P40 GPUs (24~GB VRAM each).}
\label{tab:models}
\end{table}

\paragraph{Prompts.}
Prompts are organized into two categories. \emph{Semantic prompts} consist of four domains: Wikipedia excerpts, news articles, fiction passages, and code snippets. \emph{Neutral prompts} consist of five categories designed to minimize semantic content: empty strings, random ASCII characters, explicit randomness instructions (e.g., ``Generate a completely random sequence of tokens''), neutral stubs (e.g., ``The''), and nonsense syllables (e.g., ``bla mup ziq fon''). Together, these form nine prompt categories. Each prompt-temperature combination produces a single generation of approximately 500 tokens with full logit vectors recorded at every position.

\paragraph{Temperatures.}
Three temperature values are used: 0.7, 1.0, and 1.3, spanning the range commonly used in practice.

\paragraph{Scale.}
The pilot study comprises 7,200 generations across three models and semantic prompts only. The main study adds 16,200 transformer generations (semantic and neutral) and 5,400 Mamba2 generations. The multilingual extension adds 2,400 generations with Qwen-32B across five languages (English, Japanese, Chinese, Polish, Arabic) using Wikipedia prompts. The total is 31,200 generations.

\subsection{Falsification tests}

Eight pre-registered falsification tests (F1--F8) guard against artifacts. F1 tests whether ED is significantly nonzero (one-sample $t$-test). F2 tests whether prompt domains produce different ED values (Kruskal-Wallis with Tukey HSD post-hoc). F3 tests for a model size effect (regression on $\log(\text{params})$). F4 tests temperature sensitivity (ANOVA per model). F5 tests for temporal autocorrelation in ED sequences (Durbin-Watson). F6 tests the intrinsic fraction by comparing semantic and neutral means. F7 tests the multilingual gradient (Kruskal-Wallis). F8 tests for generation-order drift (linear regression on generation index with temperature as covariate). Full results for all tests appear in Appendix~\ref{app:falsification}.

\section{Results}
\label{sec:results}

\subsection{ED is nonzero under all conditions}

Across all 31,200 generations, no model under any condition produces ED close to zero. The lowest observed mean ED is 0.238 (Llama-3 8B, pilot, semantic prompts). Even under the most ``neutral'' conditions tested (random ASCII characters), transformers produce ED $\approx 0.27$ and Mamba2 produces ED $\approx 0.54$. The one-sample $t$-test against zero (F1) returns $p < 10^{-6}$ in every partition of the data. This result is expected for any trained neural network, but the magnitude of ED and its stability across conditions are informative.

\subsection{ED is predominantly intrinsic}
\label{sec:intrinsic}

\begin{figure}[t]
\centering
\includegraphics[width=\linewidth]{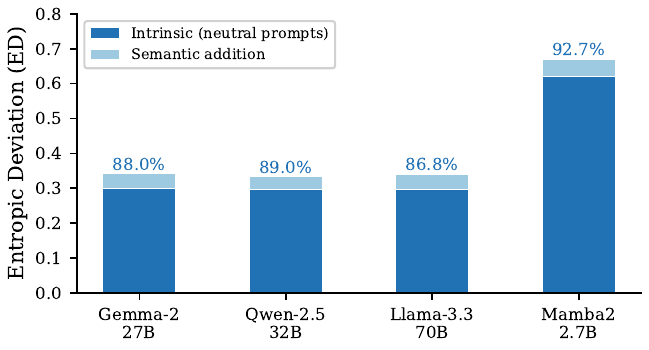}
\caption{Decomposition of ED into intrinsic (neutral prompt) and semantic components. Percentages indicate the intrinsic fraction. Across transformer families, 87--89\% of observed ED persists under semantically empty inputs. For Mamba2, the fraction reaches 93\%.}
\label{fig:intrinsic}
\end{figure}

The critical test of whether ED reflects an intrinsic structural property rather than a response to input semantics comes from comparing semantic and neutral prompts. Figure~\ref{fig:intrinsic} shows the decomposition.

For the three main-study transformers, the intrinsic fraction (neutral ED as a proportion of semantic ED) ranges from 86.8\% (Llama-3.3 70B) to 89.0\% (Qwen-2.5 32B). Mamba2 shows the highest intrinsic fraction at 92.6\%. The semantic content of a prompt adds only 7--13\% additional structure on top of a large baseline that is already present in the model's weights.

Within the neutral prompt categories, a gradient emerges. Random ASCII characters produce the lowest ED (0.268 for transformers), followed by nonsense syllables (0.286), with explicit randomness instructions (0.308), neutral stubs (0.312), and empty strings (0.314) producing higher values. The difference between ``random'' and ``empty'' is significant across all three transformers ($p < 0.001$, paired $t$-test, $t = 8.25$--$11.09$). A model explicitly instructed to ``be random'' produces slightly less concentrated distributions than a model given no input at all, suggesting that even the concept of randomness provides a minimal deconcentrating signal.

\subsection{Transformer families converge}
\label{sec:convergence}

\begin{figure}[t]
\centering
\includegraphics[width=\linewidth]{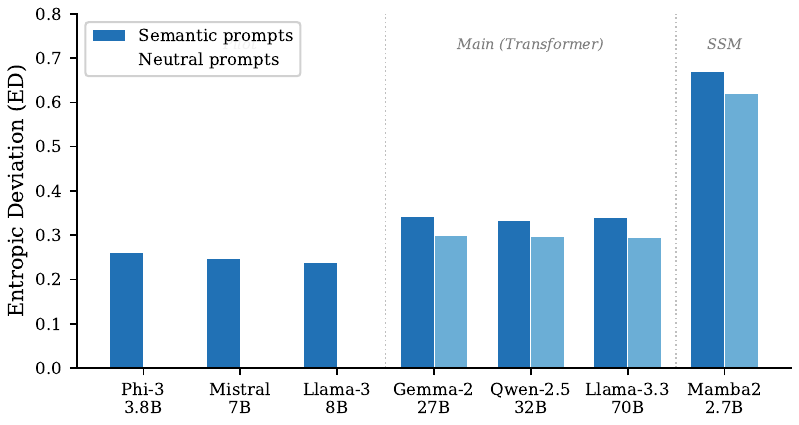}
\caption{ED across all models. Pilot models (left) show lower ED at smaller scales. The three main transformer families (center) converge on nearly identical values despite different training data, architectures, and vocabulary sizes. Mamba2 (right, SSM architecture) exhibits ED approximately twice that of transformers.}
\label{fig:models}
\end{figure}

The three main transformer families produce remarkably similar ED values despite substantial differences in training data, model architecture details, and vocabulary size (128K--256K tokens). Under semantic prompts, mean ED ranges from 0.333 (Qwen) to 0.341 (Gemma), a spread of only 0.008. Under neutral prompts, the range narrows further: 0.295 (Llama) to 0.300 (Gemma), a spread of 0.005.

More strikingly, the pattern of ED \emph{across} domains is nearly identical. The Spearman rank correlations between all pairs of transformer domain-level ED profiles range from $\rho = 0.93$ to $\rho = 0.98$ ($p < 0.001$). All three models find the same domains easier or harder in the same order, with code consistently producing the highest ED among semantic domains (0.356--0.361, significantly above other domains at $p < 10^{-39}$).

\subsection{Architecture changes the regime}
\label{sec:architecture}

Mamba2-2.7B, despite having fewer parameters than any main-study transformer, exhibits fundamentally different ED characteristics (Figure~\ref{fig:models}, right):

\paragraph{Higher baseline.} Mamba2 ED under semantic prompts averages 0.670, approximately twice the transformer mean of 0.338. Under neutral prompts, the ratio is similar: 0.620 vs. 0.297.

\paragraph{Lower variance.} Mamba2's within-sequence ED standard deviation averages 0.15, compared to 0.44 for transformers. The SSM produces more uniformly concentrated distributions across token positions.

\begin{figure}[t]
\centering
\includegraphics[width=\linewidth]{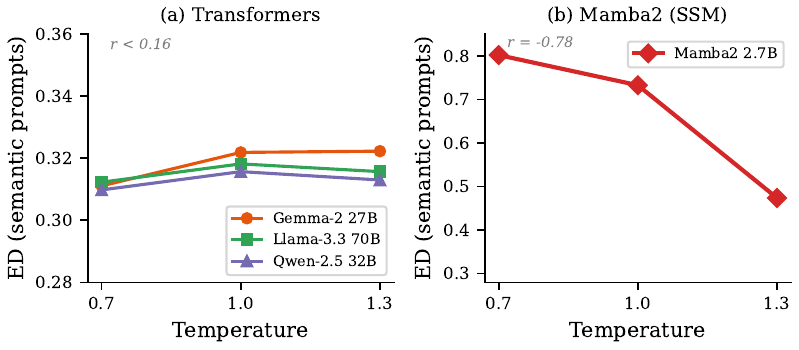}
\caption{Temperature sensitivity of ED. (a) Transformers show near-flat ED across temperatures, with a slight positive trend under semantic prompts. (b) Mamba2 shows a steep decline in ED with increasing temperature ($r = -0.78$). Note the different $y$-axis scales.}
\label{fig:temperature}
\end{figure}

\paragraph{Reversed temperature dynamics.}
Figure~\ref{fig:temperature} illustrates the most striking architectural difference. For transformers, temperature has negligible effect on ED under neutral prompts (no significant differences by Tukey HSD for Gemma; $\Delta < 0.005$ for Llama and Qwen at extreme temperatures only). Under semantic prompts, higher temperature slightly \emph{increases} transformer ED (+1.2--2.6\%), opposite to the naive expectation that higher temperature should flatten distributions. For Mamba2, temperature has a massive negative effect: ED drops from 0.796 at $T = 0.7$ to 0.440 at $T = 1.3$, a 44.7\% decrease ($r = -0.78$, all pairwise comparisons significant at $p \approx 0$ by Tukey HSD). Temperature scaling, which operates identically on the logits of both architectures, produces fundamentally different outcomes depending on how those logits were generated.

\paragraph{Partial domain agreement.}
Mamba2's domain-level ED profile correlates with transformer profiles at $\rho = 0.58$--$0.72$, substantially lower than the transformer-to-transformer correlations of $\rho > 0.93$. The architectures partially agree on which domains are ``harder'' but diverge considerably in their rankings.

\subsection{Language modulates ED independently of tokenization}
\label{sec:multilingual}

\begin{figure}[t]
\centering
\includegraphics[width=0.75\linewidth]{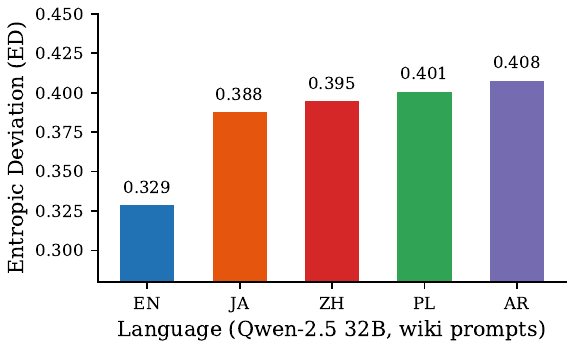}
\caption{Multilingual ED gradient for Qwen-2.5 32B on Wikipedia prompts. All pairwise differences are significant ($p < 10^{-6}$). The gradient does not track token fertility or vocabulary allocation.}
\label{fig:multilingual}
\end{figure}

Cross-lingual experiments with Qwen-2.5 32B on Wikipedia prompts in five languages reveal a stable gradient (Figure~\ref{fig:multilingual}):

\begin{equation}
\text{EN}\ (0.329) < \text{JA}\ (0.388) < \text{ZH}\ (0.395) < \text{PL}\ (0.401) < \text{AR}\ (0.408)
\end{equation}

\noindent All pairwise differences are significant (Mann-Whitney $U$, $p < 10^{-6}$) with large effect sizes (Cohen's $d = 2.4$--$4.1$ relative to English). Temperature has minimal influence on the gradient; the only significant pairwise temperature comparison across languages is $T = 0.7$ vs. $T = 1.3$ at $p = 0.03$.

The obvious hypothesis is that this gradient reflects tokenizer properties: languages requiring more tokens per character (higher fertility) might mechanically produce different ED values. Three tests reject this hypothesis:

\paragraph{Token fertility does not predict ED.} The Spearman correlation between token fertility (tokens per source character) and ED is $\rho = 0.10$, $p = 0.87$. Japanese and Chinese have more than three times the fertility of English but lower ED than Polish and Arabic.

\paragraph{Identical tokenizer, different ED.} English and Polish share an identical allocation of 104,969 tokens (69.2\% of the vocabulary) in Qwen's tokenizer, using the same Latin script subset. Despite this shared tokenizer substrate, their ED values differ by 0.072, the second-largest gap in the gradient.

\paragraph{Vocabulary usage anticorrelates with ED.} The number of unique tokens actually used in generations correlates negatively with ED ($\rho = -0.30$, $p = 0.624$). Japanese, which uses the fewest unique tokens (1,358 on average), has the lowest non-English ED. If vocabulary compression drove ED, the correlation would be positive.

After residualizing ED on estimated prompt token count to control for any length effects, the language contrasts \emph{increase} (Cohen's $d$ rises from 2.4 to 3.7, from 3.1 to 4.5, and from 4.1 to 5.2 for the three largest contrasts). The linguistic signal strengthens after controlling for tokenization artifacts.

These results support what may be called the \emph{deeper structure hypothesis}: the language in which a model operates modulates its distributional concentration through properties of the language itself (morphological complexity, syntactic predictability, $n$-gram entropy) rather than through the mechanics of how text is segmented into tokens.

\subsection{Model size effects}

The pilot models (3.8--8B parameters) produce mean ED of 0.238--0.260 under semantic prompts, substantially below the main-study transformers (27--70B, ED $= 0.313$--0.318). A regression of ED on log(parameters) across all six transformers yields $R^2 = 0.276$ ($p \approx 0$), indicating a significant but moderate size effect. Notably, within the pilot group, ED does not increase monotonically with size: Phi-3 (3.8B, ED $= 0.260$) exceeds both Mistral-7B (0.248) and Llama-8B (0.238). The relationship between model size and ED appears to involve a step between the small and large model classes rather than a smooth gradient.

The finding that Mamba2 at 2.7B parameters produces ED approximately twice that of a 70B transformer reinforces the conclusion from Section~\ref{sec:architecture}: architecture dominates over scale in determining distributional concentration.

\section{Discussion}
\label{sec:discussion}

\subsection{What the randomness floor represents}

The term ``randomness floor'' is chosen deliberately. ED under neutral prompts represents a lower bound on how random a model's token distributions can be. No amount of prompt engineering, no instruction to ``be random,'' drives ED to zero. The floor is a property of the weights, not the input.

This is not surprising in principle. Any neural network trained on non-uniform data will develop non-uniform output distributions. What is informative is the \emph{magnitude} of the effect (ED $\approx 0.30$ for transformers means the model's entropy deficit relative to uniform is 30\% of its maximum possible value), its \emph{stability} across independently trained models, and its sensitivity to architectural choices but not to temperature.

\subsection{Relationship to Zipf's law}

Natural language token frequencies follow a Zipfian distribution~\citep{zipf1949human, piantadosi2014zipf}, and models trained on such data should reflect this structure. A natural question is whether the observed ED simply recapitulates the training distribution's inherent non-uniformity. If one computes the ED of a Zipfian distribution with exponent $\alpha \approx 1.0$ (typical of English) over a vocabulary of $V = 150{,}000$ tokens, the resulting value provides a theoretical baseline. If the measured model ED exceeds this baseline, the model has learned concentration patterns beyond what raw corpus frequencies would predict. If it falls below, the model's distributional smoothing partially counteracts the Zipfian structure of its training data.

This comparison is not performed in the present work, but it constitutes a direct, falsifiable test of whether the randomness floor reflects something beyond corpus statistics. The convergence of three independently trained transformers on ED $\approx 0.31$ is suggestive: if the floor were purely a function of training data distributions, models trained on different corpora with different preprocessing would be expected to diverge more than they do.

\subsection{Relationship to perplexity}

ED and perplexity are mathematically related. Since $\ED = 1 - H(p) / \log V$ and perplexity is $\exp(H(p))$, lower perplexity (better next-token prediction) implies higher ED (greater distributional concentration). However, perplexity is typically measured on held-out text where the model has meaningful context, while the intrinsic ED is measured under conditions of minimal context. A model could have excellent perplexity on natural text while having modest intrinsic ED, if its concentration is context-dependent. Conversely, a model with high intrinsic ED would appear ``confident'' even in the absence of grounds for confidence. Comparing intrinsic ED with perplexity across models may reveal whether distributional concentration scales with prediction quality or represents an independent structural property.

\subsection{Implications for understanding model internals}

The convergence of three transformer families on nearly identical ED values, including their domain-level profiles ($\rho > 0.93$), suggests that the randomness floor may be a property of the transformer architecture and the statistical structure of natural language rather than of any particular training corpus or procedure. This convergence occurs despite vocabulary sizes ranging from 128K to 256K tokens, confirming that the normalization by $\log V$ in Equation~\ref{eq:ed} effectively controls for vocabulary differences.

The qualitative difference between transformers and Mamba2, particularly the reversed temperature sensitivity, points to a fundamental architectural distinction. In a transformer, the attention mechanism produces logit distributions that are already sharply concentrated; temperature scaling operates on distributions that have already ``decided'' their shape through attention. In an SSM, the recurrent state dynamics produce logits through a different computational pathway, one that appears more sensitive to the flattening effect of temperature division. Understanding why these architectures respond so differently to the same scaling operation may illuminate their respective computational strategies.

\subsection{Implications for sampling and decoding}

If the model's baseline distribution is already far from uniform (ED $\approx 0.30$), then common decoding strategies operate on a distribution that is pre-concentrated before any context is provided. Top-$k$ and nucleus (top-$p$) sampling~\citep{holtzman2020curious} truncate distributions that are already sharp; temperature scaling modulates something that is already far from flat. This suggests that the effective diversity of sampled outputs is bounded not only by the decoding strategy but also by the intrinsic concentration of the model's distributional prior. Characterizing this prior may help explain why certain decoding hyperparameters produce qualitatively similar outputs across wide parameter ranges.

\subsection{The multilingual gradient}

The finding that language modulates ED independently of tokenization raises questions about what linguistic properties drive the effect. English, with relatively fixed word order and limited morphology, produces the lowest ED. Arabic, with rich morphology, root-and-pattern word formation, and flexible word order, produces the highest. This pattern is consistent with the hypothesis that languages with higher structural uncertainty per token position produce higher distributional concentration: the model has stronger ``opinions'' about which tokens are likely because the space of grammatically valid continuations is more constrained by morphological agreement, case marking, or other language-specific features.

This interpretation is speculative and would require targeted experiments to confirm. What the data do establish is that the effect is real, large (Cohen's $d > 2$ for all contrasts with English), and not reducible to tokenizer artifacts.

\subsection{Threats to validity}

\paragraph{Instruction tuning confound.}
All main-study models are instruction-tuned variants (Gemma-2-IT, Qwen-2.5-Instruct, Llama-3.3-Instruct). RLHF and instruction fine-tuning reshape logit distributions to produce helpful, harmless outputs. The observed randomness floor may therefore reflect the combined effect of pretraining \emph{and} alignment, rather than pretraining alone. Base (non-instruct) models may exhibit different ED profiles. This confound applies to the absolute values of ED; it is less likely to affect cross-model comparisons (all three transformers underwent different fine-tuning procedures yet converged) or the architectural contrast with Mamba2 (which is a base model).

\paragraph{Training data overlap.}
The convergence of three transformer families on ED $\approx 0.31$ could reflect shared training data (Common Crawl, Wikipedia, and similar web corpora appear in most large-scale training sets) rather than architectural universality. Disentangling data from architecture would require models trained on fundamentally different corpora, which is rarely feasible with publicly available models.

\paragraph{Single SSM.}
The claim that SSMs exhibit a qualitatively different ED regime rests on a single model (Mamba2-2.7B). While the effect size is large (2$\times$ the transformer ED), generalization to all SSMs is premature. Testing additional SSM architectures (RWKV, Mamba-1, linear attention variants) and Mamba2 at other scales is necessary before drawing architectural conclusions.

\paragraph{Autoregressive generation path.}
ED is measured over sequences generated autoregressively: the sampled token at position $t$ becomes context for position $t+1$. The measured ED therefore depends not only on the model but also on the particular generation path. Different random seeds at the same temperature would produce different token sequences and potentially different per-position ED values, though the mean ED over many positions is expected to be stable.

\paragraph{BOS and system prompt contamination.}
The ``empty string'' neutral prompt is not truly empty. The tokenizer prepends a beginning-of-sequence (BOS) token, and instruction-tuned models may implicitly condition on a system prompt template even when none is provided. The model's response to an ``empty'' input is thus a response to minimal but nonzero context. The random ASCII and nonsense syllable conditions partially address this concern by providing context that is present but semantically vacuous.

\subsection{Limitations}

Beyond the threats above, several practical limitations apply. The multilingual analysis uses a single model (Qwen-2.5 32B); testing additional models would strengthen the cross-lingual findings. All models were run in Q4\_K\_M quantization, which may introduce small distributional shifts relative to full-precision inference. The temperature range (0.7--1.3) covers common practice but excludes extremes that might reveal nonlinear behavior. Finally, the study reports ED as a mean over token positions, which may obscure position-dependent patterns.

\section{Future work}
\label{sec:future}

Several directions follow directly from the present findings. First, comparing base and instruction-tuned variants of the same model family (e.g., Llama-3.3 base vs. Instruct) would isolate the contribution of alignment to the randomness floor. Second, computing the ED of empirical token frequency distributions from training corpora would establish a Zipfian baseline against which model ED can be compared. Third, testing additional SSM architectures (RWKV, Mamba-1, RetNet) and Mamba2 at larger scales would clarify whether the high-ED, temperature-sensitive regime is specific to Mamba2 or characteristic of recurrent architectures generally. Fourth, extending the temperature range to extremes ($T < 0.3$, $T > 2.0$) may reveal phase transitions or saturation effects in ED. Fifth, a positional analysis of ED (how concentration varies across token positions within a generation) could reveal whether the randomness floor is uniform or exhibits structure related to sequence position, attention patterns, or generation length. Finally, the relationship between intrinsic ED and standard evaluation metrics (perplexity, downstream task performance) deserves systematic investigation.

\section{Conclusion}
\label{sec:conclusion}

This paper has introduced Entropic Deviation as a metric for measuring how far language model token distributions are from uniform, and applied it systematically across models, architectures, prompts, temperatures, and languages. The findings establish that language models maintain a substantial ``randomness floor'': a level of distributional concentration that persists even under conditions designed to minimize contextual influence. For transformers, this floor is remarkably consistent across independently trained families and nearly immune to temperature. For state space models, the floor is higher and strongly temperature-dependent, revealing a qualitative architectural difference. Across languages, the floor varies in a pattern that tracks linguistic properties rather than tokenizer mechanics.

These results provide a new angle on what language models learn. The randomness floor is not a preference or a bias in the behavioral sense; it is a structural property of the learned representations, measurable at the logit level before any sampling or decoding takes place. Characterizing this property across architectures and languages may offer a complementary perspective to existing interpretability approaches that focus on individual neurons or attention patterns.

All code and data are publicly available at \url{https://github.com/JaroslawHryszko/entropic-deviation}.

\bibliography{references}
\bibliographystyle{colm2026_conference}

\appendix

\section{Falsification test results}
\label{app:falsification}

Table~\ref{tab:falsification} summarizes the outcomes of all pre-registered falsification tests across datasets.

\begin{table}[H]
\centering
\small
\begin{tabular}{lll}
\toprule
\textbf{Test} & \textbf{Description} & \textbf{Result} \\
\midrule
F1 & ED $\neq 0$ (one-sample $t$-test) & $p < 10^{-6}$ in all partitions \\
F2 & Domain differences (Kruskal-Wallis) & Significant in all models \\
F3 & Size effect (regression on $\log$ params) & $R^2 = 0.276$, $p \approx 0$ \\
F4 & Temperature effect (ANOVA) & TF: negligible; SSM: $p \approx 0$ \\
F5 & Autocorrelation (Durbin-Watson) & Weak negative (TF neutral) \\
F6 & Intrinsic fraction (semantic vs neutral) & 86.8--92.6\% intrinsic \\
F7 & Multilingual gradient (Kruskal-Wallis) & $p = 8.4 \times 10^{-45}$ \\
F8 & Generation drift (regression on index) & Slope $\sim 1.5 \times 10^{-5}$, $p < 10^{-14}$ \\
\bottomrule
\end{tabular}
\caption{Summary of pre-registered falsification tests. F8 identified a small but statistically significant generation-order drift, likely attributable to hardware state or numerical precision changes during long experimental runs. Shuffling temperature and prompt order across runs is recommended for future work.}
\label{tab:falsification}
\end{table}

\section{Neutral prompt categories}
\label{app:neutral}

Table~\ref{tab:neutral} shows ED values for each neutral prompt category, averaged across the three main transformers and separately for Mamba2.

\begin{table}[H]
\centering
\small
\begin{tabular}{lcc}
\toprule
\textbf{Category} & \textbf{TF Mean ED} & \textbf{Mamba2 Mean ED} \\
\midrule
Random ASCII & 0.268 & 0.540 \\
Nonsense syllables & 0.286 & 0.535 \\
Explicit randomness & 0.308 & 0.665 \\
Neutral stubs & 0.312 & 0.659 \\
Empty string & 0.314 & 0.704 \\
\bottomrule
\end{tabular}
\caption{ED by neutral prompt category. TF Mean is the average across three main transformers. Note the reversed ordering for Mamba2: empty strings produce the \emph{highest} ED, consistent with the interpretation that absence of context strengthens default distributional preferences in the SSM architecture.}
\label{tab:neutral}
\end{table}

\section{Domain-level ED profiles}
\label{app:domains}

Table~\ref{tab:domains} reports ED by semantic domain across models. All values are averaged over three temperature conditions.

\begin{table}[H]
\centering
\small
\begin{tabular}{lcccc}
\toprule
\textbf{Domain} & \textbf{Gemma-2} & \textbf{Qwen-2.5} & \textbf{Llama-3.3} & \textbf{Mamba2} \\
\midrule
Code & 0.356 & 0.360 & 0.361 & 0.741 \\
News & 0.342 & 0.327 & 0.338 & 0.648 \\
Fiction & 0.339 & 0.336 & 0.337 & 0.675 \\
Wikipedia & 0.337 & 0.329 & 0.337 & 0.662 \\
\bottomrule
\end{tabular}
\caption{ED by semantic prompt domain across models. Code consistently produces the highest ED in all architectures. Mamba2 domains are limited to the four categories included in the SSM experimental runs.}
\label{tab:domains}
\end{table}

\section{Multilingual tokenizer analysis}
\label{app:tokenizer}

Table~\ref{tab:multilingual} presents ED alongside tokenizer properties for each language. The key finding is the dissociation between tokenizer-level metrics (fertility, vocabulary allocation) and ED.

\begin{table}[H]
\centering
\small
\begin{tabular}{lccccc}
\toprule
\textbf{Language} & \textbf{ED} & \textbf{Fertility} & \textbf{Vocab alloc.} & \textbf{Unique gen.} & \textbf{Cohen's $d$} \\
\midrule
English & 0.329 & 0.221 & 104,969 & 1,996 & -- \\
Japanese & 0.388 & 0.747 & 38,877 & 1,358 & 2.4 \\
Chinese & 0.395 & 0.750 & 37,115 & 1,808 & 3.1 \\
Polish & 0.401 & 0.352 & 104,969 & 1,815 & 3.7 \\
Arabic & 0.408 & 0.413 & 15,429 & 1,712 & 4.1 \\
\bottomrule
\end{tabular}
\caption{Multilingual ED with tokenizer properties. Fertility = tokens per source character. Vocab alloc. = number of tokenizer entries allocated to the language's script. Unique gen. = mean unique tokens per generation. Cohen's $d$ is relative to English. Note that English and Polish share identical vocabulary allocation yet differ by $d = 3.7$.}
\label{tab:multilingual}
\end{table}

\end{document}